\documentclass[conference]{IEEEtran}
\IEEEoverridecommandlockouts
\usepackage{cite}
\usepackage{amsmath,amssymb,amsfonts}
\usepackage{algorithmic}
\usepackage{graphicx}
\usepackage{tikz}
\usetikzlibrary{positioning}
\usepackage{textcomp}
\usepackage{xcolor}
\usepackage[strings]{underscore}

\usepackage[capitalize]{cleveref}

\def\BibTeX{{\rm B\kern-.05em{\sc i\kern-.025em b}\kern-.08em
    T\kern-.1667em\lower.7ex\hbox{E}\kern-.125emX}}

\usepackage{subcaption}

\newcommand{\X}{\mathcal{X}}

\newcommand{\Ex}{\mathbb{E}}

\newcommand{\bx}{{\bar{x}}}

\newcommand{\bc}{{\bar{c}}}
\newcommand{\tp}{{t+1}}
\newcommand{\tm}{{t-1}}
\newcommand{\pt}{{\prec t}}
\newcommand{\ptm}{{\prec t-1}}

\newcommand{\set}{{\succeq t}}

\newcommand{\hphi}{{\hat{\phi}}}
\newcommand{\hPhi}{{\hat{\Phi}}}

\newcommand{\Nat}{{\mathbb{N}}}

\newcommand*{\diff}{\mathop{\kern0pt\mathrm{d}}\!{}}

\DeclareMathOperator{\HS}{\text{H}}
\DeclareMathOperator{\KL}{\text{KL}}
\DeclareMathOperator{\ntic}{\text{ntic}}
\DeclareMathOperator{\NTIC}{\text{NTIC}}
\DeclareMathOperator{\mi}{\text{i}}
\DeclareMathOperator{\IG}{\text{IG}}

\begin{document}

\title{Non-trivial informational closure of a Bayesian hyperparameter*\\
\thanks{This work was made possible through the support of a grant from Templeton World Charity Foundation, Inc. The opinions expressed in this publication are those of the authors and do not necessarily reflect the views of Templeton World Charity Foundation, Inc.}
}

\author{\IEEEauthorblockN{1\textsuperscript{st} Martin Biehl}
\IEEEauthorblockA{
\textit{Araya Inc.}\\
Tokyo, Japan \\
martin@araya.org
}
\and
\IEEEauthorblockN{2\textsuperscript{nd} Ryota Kanai}
\IEEEauthorblockA{
\textit{Araya Inc.}\\
Tokyo, Japan 
}
}

\IEEEoverridecommandlockouts
\IEEEpubid{\makebox[\columnwidth]
{978-1-7281-2547-3/20/\$31.00~\copyright2020 IEEE \hfill} 
\hspace{\columnsep}\makebox[\columnwidth]{ }}

\maketitle

\IEEEpubidadjcol

\begin{abstract}
We investigate the non-trivial informational closure (NTIC) of a Bayesian hyperparameter inferring the underlying distribution of an identically and independently distributed finite random variable. For this we embed both the Bayesian hyperparameter updating process and the random data process into a Markov chain.
The original publication by Bertschinger et al. \cite{bertschinger_information_2006} mentioned that NTIC may be able to capture an abstract notion of modeling that is agnostic to the specific internal structure of and existence of explicit representations within the modeling process. 
The Bayesian hyperparameter is of interest since it has a well defined interpretation as a model of the data process and at the same time its dynamics can be specified without reference to this interpretation. 
On the one hand we show explicitly that the NTIC of the hyperparameter increases indefinitely over time. On the other hand we attempt to establish a connection between a quantity that is a feature of the interpretation of the hyperparameter as a model, namely the information gain, and the one-step pointwise NTIC which is a quantity that does not depend on this interpretation. We find that in general we cannot use the one-step pointwise NTIC as an indicator for information gain. 
We hope this exploratory work can lead to further rigorous studies of the relation between NTIC and modeling.

\end{abstract}

\begin{IEEEkeywords}
non-trivial informational closure, information gain, Bayesian inference, modeling, individuality, autonomy, agency
\end{IEEEkeywords}

\section{Introduction}
One of the fundamental questions in artificial life research is how to identify (artificial) life within real world or simulated data. 
Even in the case where all there is to know about a system is known, as in the case of cellular automata, there is no accepted method for identifying living structures (see \cite{beer_characterizing_2014} for an approach). 
Instead of trying to detect life directly (which would require a quantitative measure of life itself) it may be possible to quantify features of life and then combine them in order to detect life. Among the features of life that are studied in artificial life are autonomy, individuality, and cognition. A quantitative measure that has been proposed in relation to all three of these concepts is non-trivial informational closure (NTIC) \cite{bertschinger_information_2006}. 
NTIC is defined for two stochastic processes, say $\Xi_t$ and $X_t$. Roughly, $\Xi_t$ is non-trivially informationally closed from $X_t$ if we can predict something from $X_t$'s past $X_\pt$ about $\Xi_t$'s future $\Xi_\set$ but whatever we can predict we could also predict from $\Xi_t$'s own past $\Xi_\pt$.
In the original work the point of departure for NTIC were cognitive systems that ``are assumed to be capable of reducing the information flow from the environment into the system by modeling the environment'' \cite{bertschinger_information_2006}. Additionally and equally relevant to the present publication the authors considered NTIC as ``an abstract notion of `modeling' that does not depend upon the identification of certain structures in the system as explicit models or representations'' \cite{bertschinger_information_2006}.
In more recent work NTIC was considered as a measure of the difference between two types of individuality, namely colonial and organismal individuality \cite{krakauer_information_2020}. Concerning autonomy, \cite{vakhrameev_measuring_2020} propose to use NTIC as a quantification of self-control which is one of two dimensions (the other being self-organization) of autonomy. Other proposed applications of NTIC are as a utility function for unsupervised learning \cite{guttenberg_neural_2016} and as part of a formalization of consciousness \cite{chang_information_2019}. 

In this work we focus on the ideas behind the original conception of NTIC. The first contribution is the study of NTIC of a new concrete example system. In \cite{bertschinger_information_2006} an example system was studied as well. There the state transition function $F(\Xi_t,X_t)$ of the process $\Xi_t$ in response to observations $X_t$ was optimized to achieve NTIC. Here, we fix this state transition function to one that implements Bayesian inference of the process that generates its observations. This leads to a prime example of a system that should be ``capable of reducing the information flow from the environment into the system by modeling the environment'' because it indeed does infer (in the infinite data limit) the correct model of the environment in a well defined way.

There is one additional aspect of our example. As we discuss in \cite{biehl_dynamics_2020} and to a lesser degree below, the interpretation as a modeling process of $\Xi_t$ is implemented by a function that maps any $\xi_t$ to a probability distribution $q(\hPhi|\xi_t)$ and the subsequent $\xi_\tp$ to the posterior $q(\hPhi|x_t,\xi_t)$ i.e.\  $q(\hPhi|\xi_\tp)=q(\hPhi|x_t,\xi_t)$. However, the result of this function i.e.\ the distribution $q(\hPhi|\xi_t)$ is not necessary to determine $\xi_\tp$. This means that the modeling really is \textit{just} an interpretation. As mentioned in \cite{biehl_dynamics_2020} this is reminiscent of the variational densities occurring in the approximate Bayesian inference lemma in \cite{friston_free_2019}. Note that, in order to compute NTIC we also only require $\Xi_t$ itself and not the distributions associated to it. In this sense NTIC remains independent of the formally well defined interpretation in terms of a model. At the same time, we can compute quantities that are associated to the model distribution $q(\hPhi|\xi_t)$ and depend on it. Examples are the information gain and the surprise associated to an observation. We make use of these properties as we explain next. 

Our choice of system can be seen as a sanity check with respect to the question of whether NTIC is ``an abstract notion of `modeling' that does not depend upon the identification of certain structures in the system as explicit models or representations'' \cite{bertschinger_information_2006} since a system that does indeed contain an explicit model must be classified as a modeling system by such a notion.
Let us consider this aspect in a bit more detail
. In general we would like a criterion such that when given two interacting processes $\Xi_t$ and $X_t$ it tells us whether $\Xi_t$ models $X_t$ without the need for any additional information. NTIC is one candidate for such a criterion (ignoring where to put a threshold for the moment). Clearly, in cases where we know that $\Xi_t$ becomes a perfect model of $X_t$ over time any suitable criterion should indicate this. Our study confirms this for NTIC. 
We could try to go further, however, and try to use NTIC to also identify more specific properties of a modeling process. One such property is the information gain which measures the change in the belief distribution over the data generating process due to an observation. Another is the surprise which is the probability of an observation according to the belief distribution just before it is observed. One question is then whether some component of NTIC can be used as an abstract notion of information gain or surprise similar to how NTIC may be an abstract notion of modeling. Specializing NTIC can be done in two ways, by looking at the pointwise NTIC of which the standard NTIC is the expectation value, and by looking at the one-step version instead of the original notion that considers the entire past observations. We combine both specializations to see whether the one-step pointwise NTIC can be used as an indicator for information gain or surprise at least in the case of the hyperparameter process. However, our results in this respect remain inconclusive.



In \cref{sec:setting} we describe the setting formally, in \cref{sec:results} we present our results and a discussion. The technical details are in \cref{sec:details}. 

\section{Setting}
\label{sec:setting}
\begin{figure}[htbp]
\begin{center}
    
        \begin{tikzpicture}
    [->,>=stealth,auto,node distance=2cm,
    thick]
    \tikzset{
			hv/.style={to path={-| (\tikztotarget)}},
			vh/.style={to path={|- (\tikztotarget)}},
    }

    \node (e) [] {$\Xi_1$};
    \node (e') [right of=e] {$\Xi_{2}$};
    \node (s) [below of=e] {$X_1$};
    \node (s') [below of=e'] {$X_{2}$};
    \node (el) [left of=e] {$\xi_0$};
    \node (er) [right of=e'] {};
    \node (sl) [below of=el] {$X_0$};
    \node (sr) [below of=er] {};
    \node (phi) [below left=.5cm and .5cm of sl] {$\phi$};
    \path
      (e) edge node {} (e')
      (s) edge node {} (e')
      (sl) edge node {} (e)
      (el) edge node {} (e)
      (e') edge[-,dotted] node {} (er)
      (s') edge[-,dotted] node {} (er)
      (phi) edge[hv] (sl)
      (phi) edge[hv] (s)
      (phi) edge[hv] (s')
      (phi) edge[-,dotted,hv] (sr)
      
      ;
  \end{tikzpicture}
\end{center}

\caption{Bayesian network of the hyperparameter updating process for an IID process with parameter $\phi$ and initial hyperparameter $\xi_0$.}  
\label{fig:bn}%
\end{figure}
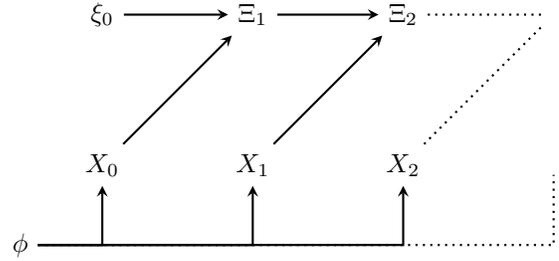%

Assume as given an identically and independently distributed (IID) random process $(X_t)_{t \in \mathbb{N}}$ with finite sample space $\X$ specified by a categorical distribution
with parameter $\phi=(\phi_x)_{x \in \X}$ which is a vector of the probabilities of the different outcomes i.e.\ $\phi_x \in [0,1]$ and 
\begin{align}
\sum_{x \in \X} \phi_x =1.
\end{align}
For each $t \in \mathbb{N}$ we then have
\begin{align}
\label{eq:iidx}
  p(x_t|\phi)=\prod_x \phi_x^{\delta_{x x_t}}
\end{align}
where $\delta_{x x_t}$ is the Kronecker delta.

Then assume another process $(\Xi_t)_{t \in \mathbb{N}}$ whose dynamics are those of a Bayesian hyperparameter (specifically a parameter of a Dirichlet distribution over parameters of categorical distributions) that updates to parameterize the posterior after each sample from $(X_t)_{t \in \mathbb{N}}$. More precisely, we imagine that for all $t \in \mathbb{N}$ the outcome $\xi_t$ parameterizes a Dirichlet distribution $q(\hphi|\xi_t)$ over possible values $\hphi$ of the true categorical distribution parameter $\phi$. After observing a new sample $x_t$ the posterior $q(\hphi|x_t,\xi_t)$ is then well defined. To update $\xi_t$ to $\xi_\tp$ we require that $\xi_\tp$ is the parameter of the posterior i.e.\:
\begin{align}
  q(\hphi|\xi_\tp)=q(\hphi|x_t,\xi_t).
\end{align}

%
Since $\xi_t$ is the parameter of a Dirichlet distribution over categorical parameters we can directly calculate $\xi_\tp$ from $\xi_t$ and $x_t$ via (see e.g.\ \cite{bishop_pattern_2006}):
\begin{align}
\label{eq:iidxiupdate}
  \xi_\tp(x_t,\xi_t)=\xi_t+\delta_{x_t}
\end{align}
since $\xi_t$ are vectors with $|\X|$ components we can also write this (maybe more clearly) componentwise, i.e. for each component $x \in \X$:
\begin{align}
  (\xi_\tp)_x:=(\xi_t)_x+(\delta_{x_t})_x
\end{align}
Here $(\delta_{x_t})_x:=\delta_{x_t x}$. In other words, $\delta_{x_t}$ is a one-hot encoding of $x_t$. For later use also note that we can combine multiple updates due to a sequence of observations $x_{t:t+n}:=(x_t,...,x_{t+n-1})$ by just adding them up:
\begin{align}
  \xi_{t+n}(x_{t:t+n},\xi_t)&=\xi_t + \sum_{\tau=t}^{t+n-1} \delta_{x_\tau}\\
  &=\xi_t + c(x_{t:t+n}).
\end{align}
Here we defined the counting function $c$ which will be used extensively: 
\begin{align}
  c(x_{t:t+n})_x:=\sum_{\tau=t}^{t+n-1} \delta_{x_\tau}.
\end{align}
In words $c(x_{t:t+n}):=(c(x_{t:t+n})_x)_{x \in \X}$ is the vector of counts of the different outcomes $x \in \X$ within the sample sequence $x_{t:t+n}$. Note that in our notation $c(x_{t:t+1})=c(x_t)=\delta_{x_t}$. The inverse $c^{-1}(\bar{c})$ of the counting function maps a given count $\bar{c}$ to all the data sequences that produce this count and will be used in the following as well. In particular we need the cardinality $|c^{-1}(\bar{c})|$ of the resulting set $c^{-1}(\bar{c})$ of sequences which is a multinomial coefficient:
\begin{align}
\label{eq:multinomial}
  |c^{-1}(\bar{c})|=\frac{(\sum_{x} \bar{c}_x)!}{\prod_x \bar{c}_x!}.
\end{align}
This is the number of trajectories that produce the count $\bar{c}$.
%

Coming back to the specification of our system, the dynamics of $\xi_t$ expressed as transition probabilities are
\begin{align}
\label{eq:iidximechanism}
  p(\xi_\tp|\xi_t,x_t):=\delta_{\xi_t+\delta_{x_t}}(\xi_\tp).
\end{align}
Together with \cref{eq:iidx} this fixes all the mechanisms/kernels in the Bayesian network \cref{fig:bn} which illustrates our setting. 
From \cite{biehl_dynamics_2020} we recall:
\begin{quote}
While we derived the dynamics of $\xi_t$ from Bayesian inference, the resulting dynamics \cref{eq:iidxiupdate} are just those of a counter of occurrences. There is no reference anymore to the belief $q(\hPhi|\xi_t)$. 
\end{quote}
Finally, note that in our calculations we always fix the parameter $\phi$ that specifies the IID process $(X_t)_{t \in \mathbb{N}}$ and the initial hyperparameter $\xi_0$. 
The latter represents knowledge contained in the hyperparameter prior to the observations $(x_t)_{t \geq 0}$ we consider.

\section{Results}
\label{sec:results}
We first present the results for the NTIC measure that considers all past observations $x_\pt$ which we call the full past version. Then we also look at the case where only the last observation is considered for the NTIC measure.
\subsection{Full past NTIC results}
The first result is the standard NTIC of the hyperparameter process $\Xi_t$ at $t$. This is defined as \cite{bertschinger_information_2006}:
\begin{align}
\NTIC_t:&= I(X_\pt:\Xi_t)-I(\Xi_t:X_\tm|\Xi_\tm).
\end{align}
Here, $I(X_\pt:\Xi_t)$ is the mutual information between the past observations $X_\pt$ and current hyperparameter $\Xi_t$ and $I(\Xi_t:X_\tm|\Xi_\tm)$ is the transfer entropy from the observations into the hyperparameter process.
So we get:
\begin{align}
\begin{split}
&\NTIC_t(\xi_0,\phi)\\
&:= I(X_\pt:\Xi_t|\xi_0,\phi)-I(\Xi_t:X_\pt|\Xi_\tm,\xi_0,\phi)\end{split}
\end{align}
As we show in \cref{sec:fpntic} this is equal to (see \cref{eq:iidNTICresult}):
\begin{align}
\label{eq:iidNTICresultpres}
 \NTIC_t(\phi)
 &=\HS(C_t|\phi)-\HS(X_\tm|\phi).
\end{align}
In words, NTIC at time $t$ is equal to the entropy of the observation count $C_t$ minus the entropy of a single observation $X_\tm$ (all observations have the same entropy in our setting). Like all other versions of NTIC we compute here it does not depend on the initial hyperparamter $\xi_0$ and therefore also cannot contain any information about it. The entropy of a single observation  corresponds to the transfer entropy term and is constant over time. The entropy of the observation count on the other hand is equal to $\HS(X_\tm|\phi)$ at $t=1$ since the count only contains the one observation $X_\tm$. For larger timesteps it grows unbounded strictly monotonously as more and more counts become possible unless the observations generated form $\phi$ have zero entropy in which case NTIC is constant and zero.
In summary the Bayesian hyperparameter process has diverging NTIC and the transfer entropy term becomes negligible but stays constant. So the process never becomes informationally closed.

The second result for the full past is the pointwise NTIC $\ntic_t$. The NTIC above is just the expectation value of the pointwise version:
\begin{align}
\begin{split}
\NTIC_t(\phi)
=\sum_{x_\pt}p(x_\pt|\phi) \ntic_t(x_\pt,\phi).
\end{split}
\end{align}
For details see \cref{sec:fpntic}
In contrast to the normal NTIC the pointwise NTIC is specific to single trajectories $x_\pt$. As seen in \cref{eq:iidnticresult} this is:
\begin{align}
  \ntic_t(x_\pt,\phi)
  &=\log \frac{p(x_\tm|\phi)}{p(c(x_\pt)|\phi)}.\label{eq:iidnticresultpres}
\end{align}
So the full past pointwise NTIC at time $t$ is the log ratio of the probability of the last observation to that of the count $c(x_\pt)$ of the considered trajectory $x_\pt$.

\subsection{One-step NTIC results}
The one-step NTIC in our setting is defined as
\begin{align}
\begin{split}
&\NTIC^1_t(\xi_0,\phi)\\
&:= I(X_\tm:\Xi_t|\xi_0,\phi)-I(\Xi_t:X_\pt|\Xi_\tm,\xi_0,\phi)\end{split}\\
&\phantom{:}=\sum_{\substack{x_\pt\\ \xi_t \\ \xi_\tm}}p(x_\pt,\xi_t,\xi_\tm|\xi_0,\phi) \ntic^1_t(\xi_t,\xi_\tm,x_\pt,\xi_0,\phi).
\end{align}
where we also gave the definition in terms of the one-step pointwise NTIC $\ntic^1_t$. For details see \cref{sec:osntic}. We focus more on the pointwise version but the standard one-step NTIC turns out equal to:
\begin{align}
  \NTIC^1_t(x_\pt,\phi)
  &=\sum_{x_\pt} p(x_\pt|\phi) \log \frac{c(x_\pt)_{x_\tm}}{t}.\label{eq:iidpwoNTICresultpres}
\end{align}
The one-step pointwise NTIC is accordingly (see \cref{eq:iidpwonticresult})
\begin{align}
  \ntic^1_t(x_\pt)
  &=\log \frac{c(x_\pt)_{x_\tm}}{|c(x_\pt)|}\label{eq:iidpwonticresultpres}.
\end{align}
Where we overloaded the cardinality notation to also mean $|\xi|=\sum_x (\xi)_x$.
This shows that the one-step pointwise NTIC neither depends on the initial hyperparameter $\xi_0$ nor on the parameter of the observation process but only on the trajectory itself. In words the one step pointwise NTIC is the logarithm of the relative frequency with which the last observation $x_\tm$ occurred in $x_\pt$. Another way to put this is to say it is the logarithm of the empirical probability of the last observation given all observations including it. Let us denote the empirical probability due to $x_\pt$ for any $x \in \X$ as
 \begin{align}
   \hat{q}_{x_\pt}(x):=\frac{c(x_\pt)_x}{|c(x_\pt)|}.
 \end{align}
Then the one-step pointwise NTIC can also be written as
\begin{align}
  \ntic^1_t(x_\pt)= \log \hat{q}_{x_\pt}(x_\tm).
\end{align}
This highlights that it contains a kind of hindsight probability since $x_\tm$ is contained in $x_\pt$ and is considered as data for this empirical distribution. 


\subsection{Marginal surprise result}
Marginal surprise is the negative logarithm of the probability $q(x_t|x_\pt,\xi_0)$ of observation $x_t$ according to the posterior predictive distribution $q(X_t|x_\pt,\xi_0)$ after the sequence of observations $x_\pt$. We call it marginal surprise because it is about an as yet unobserved value and takes in all observations made by time $t$.
\begin{align}
  - \log q(x_t|x_\pt,\xi_0)&=-\log q(x_t|\xi_t(x_\pt,\xi_0))\\
  &=- \log q(x_t|\xi_0+c(x_\pt))\\
&=- \log \frac{(\xi_0+c(x_\pt))_{x_t}}{|\xi_0+c(x_\pt)|}.\label{eq:msurpriseres}
\end{align}
Note here that the marginal surprise can also be calculated for the last previously observed value $x_\tm$ by simply looking at
\begin{align}
  - \log q(x_\tm|x_\pt,\xi_0)=-\log \frac{(\xi_0+c(x_\pt))_{x_\tm}}{|\xi_0+c(x_\pt)|}.
\end{align}
 In this case it is only the entries of the initial hyperparameter $\xi_0$ that make the marginal surprise different from the negative one-step pointwise NTIC. Similar to before we can think of the marginal surprise at time $t$ of the last observation $x_\tm$ as a kind of hindsight marginal surprise. It quantifies how much the model now would be surprised about the observation it just made if it were to observe it. 

 The negative of the one-step pointwise NTIC, can then be seen as the hindsight surprise according to the empirical distribution i.e.\ as the hindsight empirical surprise:
 \begin{align}
   -\log \hat{q}_{x_\pt}(x_\tm)=- \ntic^1_t(x_\pt).
 \end{align}

\subsection{Information gain results}
Information gain is defined as the KL-divergence between prior and posterior \cite{itti_bayesian_2006}. In our setting this is just the KL-divergence between the belief associated to $\xi_t$ and that associated to $\xi_0$ for the full past information gain or that between the beliefs associated to $\xi_t$ and $\xi_\tm$ for the one-step surprise.  
The full past information gain is not easy to interpret but the interested reader can find an expression in \cref{eq:fullpastig}. 
For the Bayesian hyperparameter process the one-step information gain can be expressed (in two ways) as (\cref{eq:iidpwosurpriseresult}):
\begin{align}
\begin{split}
  &\IG^1_t(x_\pt,\xi_0)\\
  &= - \log q(x_\tm|x_\ptm,\xi_0) + \\ 
  &\phantom{= - \log}-\Ex_{\xi_0+c(x_\pt)}[-\log q(x_\tm|\hPhi)]
\end{split}\\
  \begin{split}
&=-  \log \frac{(\xi_0)_{x_\tm}-1+c(x_\pt)_{x_\tm}}{|\xi_0|-1+|c(x_\pt)|}+\\
  &\phantom{=\Psi}+\Psi((\xi_0+c(x_\pt))_{x_\tm})-\Psi(|\xi_0+c(x_\pt)|).
\end{split}
\end{align}
Where the first expression shows that the KL divergence can be written as the difference between the marginal surprise of $x_\tm$ at time $\tm$ and the expected value according to belief $q(\hphi|\xi_0+c(x_\pt))$ at time $t$ of the (hindsight) surprise about $x_\tm$ according to model $\hphi$. 
The second expression
highlights the occurrence of $c(x_\pt)_{x_\tm}$ and $|c(x_\pt)|=t$ in the marginal surprise term and indicates that the expectation value has an analytic solution in terms of the Digamma function $\Psi$. 

The first expression is more suggestive. The KL divergence takes the marginal surprise about the last observation $x_\tm$ according to the hyperparameter $\xi_\tm$ at $t-1$ i.e.\ before taking $x_\tm$ into account. From this surprise it subtracts the expected (hindsight) surprise about the last observation $x_\tm$ over all models according to the new hyperparameter $\xi_t$ that takes into account the last observation.

It is striking that the one-step pointwise NTIC $\ntic^1_t$ at time $t$ has similarities with both expressions. On the one hand, its negative, the hindsight empirical surprise, is similar to the marginal surprise term $-\log q(x_\tm|x_\ptm,\xi_0)$. The differences come from the initial hyperparameter $\xi_0$ in the marginal surprise term and the fact that the marginal surprise term for the information gain from $t-1$ to $t$ only takes into account observations up to $t-2$ while the one-step pointwise NTIC takes into account the observations up to $t-1$.

On the other hand, the second term, the expected value term is similar because it takes the observation at $t-1$ into account and considers the probabilities of that last observation $x_\tm$ as well. Again, the expected value also takes the initial hyperparameter into account and it is an expectation value over all model parameters $\hphi$. 

Note however, that we can generally not use the one-step pointwise NTIC as an indicator for the information gain. 
For the same trajectory, the one-step pointwise NTIC will always be the same but we can change the information gain in any direction by manipulating the initial hyperparameter $\xi_0$. One, way to interpret this is that the prior experience of the hyperparameter remains undetected by the one-step pointwise NTIC and therefore the information gain cannot be identified. Similarly, the change of the one-step pointwise NTIC from time $t-1$ to $t$ does not tell us whether the information gain is high or low. Two different events that have both not been observed before, if observed in succession, have constant one-step pointwise NTIC. However, depending on the hyperparameter the information gain values can be arbitrary.

\section{Conclusion}
We have proposed a system setup that lets us study non-trivial informational closure (NTIC) of a Bayesian hyperparameter. We found that the hyperparameter process has monotonously increasing full past NTIC. This is the originally proposed NTIC measure of \cite{bertschinger_information_2006}. The hyperparameter process never becomes informationally closed since the transfer entropy is constant. 
We have also calculated the one-step NTIC, and pointwise versions of the full past and one-step NTIC. For the one-step pointwise NTIC of the hyperparameter process we found that it is equal to the logarithm of the relative frequency (or empirical probability) of the last observation within the data sequence. In an effort to relate this one-step pointwise NTIC to the information gain that the hyperparameter process encodes we highlighted connections to marginal surprise and the expected value of (hindsight) surprise. However, in general we cannot use the one-step pointwise NTIC as an indicator for information gain. 
In future work we hope to get a deeper understanding of how NTIC can be used to characterize processes that can be interpreted as having a model.

\section{Calculation details}
\label{sec:details}
\subsection{IID NTIC}
\subsubsection{Full past NTIC}
\label{sec:fpntic}
We want to compute NTIC which is defined in \cite{bertschinger_information_2006}:
\begin{align}
\label{eq:NTIC}
\NTIC_t:&= I(X_\pt:\Xi_t)-I(\Xi_t:X_\pt|\Xi_\tm)\\
&= I(X_\pt:\Xi_t)-I(\Xi_t:X_\tm|\Xi_\tm).
\end{align}
Where we used that $p(\xi_t|x_\pt,\xi_\tm)=p(\xi_t|x_\tm,\xi_\tm)$ to simplify the second term. This follows from \cref{fig:bn} via $d$-separation.
Including the initial conditions i.e.\ the initial hyperparameter $\xi_0$ 
and the parameter $\phi$ we get:
\begin{align}
\begin{split}
\label{eq:NTICparam}
&\NTIC_t(\xi_0,\phi)\\
&:= I(X_\pt:\Xi_t|\xi_0,\phi)-I(\Xi_t:X_\pt|\Xi_\tm,\xi_0,\phi)\end{split}\\
&\phantom{:}=\sum_{\substack{x_\pt\\ \xi_t \\ \xi_\tm}}p(x_\pt,\xi_t,\xi_\tm|\xi_0,\phi) \ntic_t(\xi_t,\xi_\tm,x_\pt,\xi_0,\phi).
\end{align}
Here we re-expressed NTIC as the expectation value of the pointwise non-trivial information closure $\ntic_t$. This is defined via pointwise informations measures as:
\begin{align}
\begin{split}
  \ntic_t&(\xi_t,\xi_\tm,x_\pt,\xi_0,\phi) \\
  :&=\mi(x_\pt:\xi_t|\xi_0,\phi)-\mi(\xi_t:x_\tm|\xi_\tm,\xi_0,\phi)                                                                    \end{split}
\end{align}
The pointwise mutual informations is 
\begin{align}
  \mi(x_\pt:\xi_t|\xi_0,\phi):=\log \frac{p(\xi_t|x_\pt,\xi_0,\phi)}{p(\xi_t|\xi_0,\phi)}
\end{align}
and the pointwise transfer entropy is
\begin{align}
  \mi(\xi_t:x_\tm|\xi_\tm,\xi_0,\phi):=\log\frac{p(x_\tm|\xi_t,\xi_\tm,\xi_0,\phi)}{p(x_\tm|\xi_\tm,\xi_0,\phi)}.
\end{align}
Next note that both $\xi_t$ and $\xi_\tm$ are uniquely determined by $x_\pt$ and $\xi_0$ together via the counting function:
\begin{align}
  \xi_t(x_\pt,\xi_0)&=\xi_0+c(x_{\pt})\\
    \xi_\tm(x_{\pt-1},\xi_0)&=\xi_0+c(x_{\pt-1})
\end{align}
such that these are the only values of $\xi_t,\xi_\tm$ that can occur in any run of this system:
\begin{align}
  \begin{split}
p(&x_\pt,\xi_t,\xi_\tm|\xi_0,\phi)\\
  &=p(\xi_t,\xi_\tm|x_\pt,\xi_0,\phi)p(x_\pt|\xi_0,\phi)
\end{split}\\
  &=\delta_{\xi_0+c(x_{\pt-1})}(\xi_\tm)\delta_{\xi_\tm+c(x_\tm)}(\xi_t)p(x_\pt|\xi_0,\phi).
\end{align}
So the only values of $\ntic_t$ that can occur are those with these particular values of $(\xi_t,\xi_\tm)$. 
Without loss of generality we can therefore restrict ourselves to these cases and define:
\begin{align}
  \begin{split}
\ntic_t&(x_\pt,\xi_0,\phi)\\
&:=\ntic_t(\xi_t(x_\pt,\xi_0),\xi_\tm(x_{\pt-1},\xi_0),x_\pt,\xi_0,\phi).
\end{split}
\end{align}
Then simplify:
\begin{align}
\label{eq:iidNTIC2}
&\NTIC_t(\xi_0,\phi)
=\sum_{x_\pt}p(x_\pt|\phi) \ntic_t(x_\pt,\xi_0,\phi).
\end{align}
We now calculate $\ntic_t(x_\pt,\xi_0,\phi)$. First, the mutual information term with $\xi_t(x_\pt,\xi_0)$:
\begin{align}
\label{eq:ontrapwmi}
\begin{split}
  \mi(&x_\pt:\xi_t(x_\pt,\xi_0)|\xi_0,\phi)\\
  &=\log \frac{p(\xi_t(x_\pt,\xi_0)|x_\pt,\xi_0,\phi)}{p(\xi_t(x_\pt,\xi_0)|\xi_0,\phi)}                                                                                        \end{split}
\\
  &=\log \frac{1}{p(\xi_t(x_\pt,\xi_0)|\xi_0,\phi)}.
\end{align}
Then the transfer entropy term:
\begin{align}
  \begin{split}
\mi(&\xi_t(x_\pt,\xi_0):x_\tm|\xi_\tm(x_\pt,\xi_0),\xi_0,\phi)\\
  &=\log\frac{p(x_\tm|\xi_t(x_\pt,\xi_0),\xi_\tm(x_\pt,\xi_0),\xi_0,\phi)}{p(x_\tm|\xi_\tm(x_\pt,\xi_0),\xi_0,\phi)}                                                                                                                    \end{split}
\\
  &=\log\frac{1}{p(x_\tm|\phi)} \label{eq:ontrapwte}
\end{align}
Where we used that knowing $\xi_t$ and $\xi_\tm$ determines uniquely which $\bx_\tm \in \X$ must have occurred so that the probability of any $\bx_\tm$ is either $0$ or $1$. Since $x_\tm$ is part of $x_\pt$ we assume it has occurred which means it cannot have probability $0$. 
This results in the numerator simplifying to $1$. For the denominator we used $d$-separation according to which we have
\begin{align}
  p(x_\tm|\xi_\tm,\xi_0,\phi)=p(x_\tm|\phi).
\end{align}
So together we get:
\begin{align}
  \ntic_t(x_\pt,\xi_0,\phi)=\log \frac{p(x_\tm|\phi)}{p(\xi_t(x_\pt,\xi_0)|\xi_0,\phi)}.
\end{align}
Next, we take a closer look at the denominator in the fraction. For a general $\xi_t$ this is:
\begin{align}
  p(\xi_t|\xi_0,\phi)&= \sum_{\bx_\pt} p(\xi_t,\bx_\pt|\xi_0,\phi)\\
                     &= \sum_{\bx_\pt} p(\xi_t|\bx_\pt,\xi_0) p(\bx_\pt|\phi)\\
                     &= \sum_{\bx_\pt} \delta_{c^{-1}(\xi_t-\xi_0)}(\bx_\pt) p(\bx_\pt|\phi)\\
                     &= \sum_{\bx_\pt \in c^{-1}(\xi_t-\xi_0)}  p(\bx_\pt|\phi)\\ 
                     &= p(\xi_t-\xi_0|\phi)  \label{eq:xi0toxit}                       
\end{align}
Here used a slightly generalized Kronecker-delta notation. For any set $A$:
\begin{align}
  \delta_{A}(x):=\begin{cases}
                   1 &\text{ if } x \in A\\
                   0 &\text{ else.}
                 \end{cases}
\end{align}
Now note that for the particular $\xi_t$ generated by data sequence $x_\pt$ and initial hyperparameter $\xi_0$ i.e.\ for $\xi_t=\xi_t(x_\pt,\xi_0)$ we have
\begin{align}
  \xi_t-\xi_0&=\xi_t(x_\pt,\xi_0)-\xi_0\\
  &=c(x_\pt).
\end{align}
With this we can write:
\begin{align}
\label{eq:iidmidenum}
  p(\xi_t(x_\pt,\xi_0)|\xi_0,\phi)&= p(\xi_t-\xi_0|\phi) \\
  &=p(c(x_\pt)|\phi). 
\end{align}
This is also independent of $\xi_0$ which means that $\ntic_t(x_\pt,\xi_0,\phi)$ is independent of $\xi_0$:
\begin{align}
  \ntic_t(x_\pt,\xi_0,\phi)&=\ntic_t(x_\pt,\phi)\\
  &=\log \frac{p(x_\tm|\phi)}{p(c(x_\pt)|\phi)}.\label{eq:iidnticresult}
\end{align}
This is our main result for the pointwise full past NTIC.
Similarly, $\NTIC_t$ is independent of $\xi_0$:
\begin{align}
  \NTIC_t(\xi_0,\phi)&=\NTIC_t(\phi)\\
  &=\sum_{x_\pt} p(x_\pt|\phi) \ntic_t(x_\pt,\phi).
\end{align}
If we split up $\ntic_t$ again we can resolve the mutual information and transfer entropy term. First, we look at the mutual information term. Let $\bc_t$ stand for a count of a data sequence of length $t$ i.e.\ 
\begin{align}
\bc_t \in \{d \in \Nat^{\X}:\exists x_\pt \in \X^t, c(x_\pt)= d\}.
\end{align}
Then
\begin{align}
  \begin{split}
I(X_\pt&:\Xi_t|\xi_0,\phi)\\
&= \sum_{x_\pt} p(x_\pt|\phi)  \mi(x_\pt:\xi_t(x_\pt,\xi_0)|\xi_0,\phi)\end{split}\\
&=\sum_{\bc_t} \sum_{x_\pt \in c^{-1}(\bc_t)} p(x_\pt|\phi) \log \frac{1}{p(c(x_\pt)|\phi)}
  \\                                                
  &=\sum_{\bc_t} \log \frac{1}{p(\bc_t|\phi)} \sum_{x_\pt \in c^{-1}(\bc_t)} p(x_\pt|\phi) \\
  &=\sum_{\bc_t} \log \frac{1}{p(\bc_t|\phi)} p(\bc_t|\phi) \\  
  &=\HS(C_t|\phi).
\end{align}
So the mutual information term is equal to the entropy of the counts of data sequences with length $t$.
For the transfer entropy we get 
\begin{align}
\begin{split}
&I(\Xi_t:X_\pt|\Xi_\tm,\xi_0,\phi)\\
&=\sum_{x_\pt} p(x_\pt|\phi) \mi(\xi_t(x_\pt,\xi_0):x_\tm|\xi_\tm(x_\pt,\xi_0),\xi_0,\phi)                                                                                          \end{split}\\
&=\sum_{x_\tm} p(x_\tm|\phi)\log\frac{1}{p(x_\tm|\phi)}\\
&=\HS(X_\tm|\phi).
\end{align}
So that we get for $\NTIC_t$
\begin{align}
\label{eq:iidNTICresult}
  \NTIC_t(\phi)=\HS(C_t|\phi)-\HS(X_\tm|\phi).
\end{align}

\subsubsection{One-step NTIC}
\label{sec:osntic}
In addition to the non-trivial informational closure with respect to the whole full past data sequence $x_\pt$ we also compute the non-trivial informational closure with respect tot only the last datum $x_\tm$. This is defined as:
\begin{align}
\NTIC^1_t:&= I(X_\tm:\Xi_t)-I(\Xi_t:X_\tm|\Xi_\tm).
\end{align}
Where we can see that the second term (the transfer entropy) is the same as in the full past $\NTIC_t$ (see \cref{eq:NTIC}). We can therefore focus only on the mutual information term here.
Include the initial conditions we get:
\begin{align}
\begin{split}
\label{eq:iidontic}
&\NTIC^1_t(\xi_0,\phi)\\
&:= I(X_\tm:\Xi_t|\xi_0,\phi)-I(\Xi_t:X_\pt|\Xi_\tm,\xi_0,\phi)\end{split}\\
&\phantom{:}=\sum_{\substack{x_\pt\\ \xi_t \\ \xi_\tm}}p(x_\pt,\xi_t,\xi_\tm|\xi_0,\phi) \ntic^1_t(\xi_t,\xi_\tm,x_\pt,\xi_0,\phi).
\end{align}
We here again restrict ourselves without loss of generality to values of $\xi_t,\xi_\tm$ and therefore of $\ntic^1_t$ that can occur due to a data sequence $x_\pt$. This rules out for example $\xi_\tm$ values that cannot change into $\xi_t$ by any $x_\tm$ and also those that cannot occur under $\xi_0$. 
Without loss of generality we can therefore restrict ourselves to these cases and define:
\begin{align}
  \begin{split}
\ntic^1_t&(x_\pt,\xi_0,\phi)\\
&:=\ntic^1_t(\xi_t(x_\pt,\xi_0),\xi_\tm(x_{\pt-1},\xi_0),x_\tm,\xi_0,\phi).
\end{split}
\end{align}
Then simplify:
\begin{align}
\label{eq:iidoNTIC2}
&\NTIC^1_t(\xi_0,\phi)
=\sum_{x_\pt}p(x_\pt|\phi) \ntic^1_t(x_\pt,\xi_0,\phi).
\end{align}
We now calculate $\ntic^1_t(x_\pt,\xi_0,\phi)$. The transfer entropy term stays the same so we only need to calculate the mutual information term with $\xi_t(x_\pt,\xi_0)$:
\begin{align}
 \begin{split}
 \mi(x_\tm:\xi_t(x_\pt,\xi_0)&|\xi_0,\phi)\\
  &=\log \frac{p(\xi_t(x_\pt,\xi_0)|x_\tm,\xi_0,\phi)}{p(\xi_t(x_\pt,\xi_0)|\xi_0,\phi)}.
\end{split}
\end{align}
Note that the denominator is the same as in the full past case of \cref{eq:ontrapwmi}. However, unlike in the full past case the numerator does not trivially become $1$ so let us focus on it (dropping the dependence of $\xi_t$ on $(x_\pt,\xi_0)$ for the moment since this is just an additional assumption that we can reintroduce later):
\begin{align}
\begin{split}
  p&(\xi_t|x_\tm,\xi_0,\phi)\\
  &=\frac{1}{p(x_\tm|\xi_0,\phi)}p(\xi_t,x_\tm|\xi_0,\phi)                                                          \end{split}
\\
  &=\frac{1}{p(x_\tm|\phi)}\sum_{\xi_\tm}p(\xi_t|x_\tm,\xi_\tm)p(\xi_\tm|\xi_0,\phi)p(x_\tm|\phi)\\
  &=\sum_{\xi_\tm}p(\xi_t|x_\tm,\xi_\tm)p(\xi_\tm|\xi_0,\phi)\\  
  &=\sum_{\xi_\tm}\delta_{\xi_t-\delta_{x_\tm}}(\xi_\tm)p(\xi_\tm|\xi_0,\phi)\\
  &=p(\xi_t-\delta_{x_\tm}|\xi_0,\phi)\\
  &=p(\xi_t-\delta_{x_\tm}-\xi_0|\phi)
\end{align}
Where we used \cref{eq:xi0toxit} in the last line. We can now reintroduce $\xi_t=\xi_t(x_\pt,\xi_0)$ to get
\begin{align}
  p&(\xi_t|x_\tm,\xi_0,\phi)=p(c(x_\ptm)|\phi).
\end{align}
Combined with the result for the denominator of \cref{eq:iidmidenum} we get:
\begin{align}
   \begin{split}
\mi(x_\tm&:\xi_t(x_\pt,\xi_0)|\xi_0,\phi)\\
   &=\log \frac{p(c(x_\ptm)|\phi)}{p(c(x_\pt)|\phi)}                                                    \end{split}
\\
&=\log \frac{p(x_\ptm|\phi) |c^{-1}(c(x_\ptm)|}{p(x_\pt|\phi) |c^{-1}(c(x_\pt)|}\\
&=\log \frac{\frac{(\sum_x c(x_\ptm)_x)!}{\prod_x c(x_\ptm)_x!}}{p(x_\tm|\phi) \frac{(\sum_x c(x_\pt)_x)!}{\prod_x c(x_\pt)_x!}}\\
&=\log \frac{\frac{(t-1)!}{(c(x_\pt)_{x_\tm}-1)! \prod_{x\neq x_\tm} c(x_\pt)_x!}}{p(x_\tm|\phi) \frac{t!}{c(x_\pt)_{x_\tm}! \prod_{x\neq x_\tm} c(x_\pt)_x!}}\\
&=\log \frac{c(x_\pt)_{x_\tm}}{t \,p(x_\tm|\phi) }
\end{align}
Where we used $\sum_x c(x_\pt)_x=t$ and
\begin{align}
  p(c(x_\pt)|\phi)&=\sum_{\bx_\pt \in c^{-1}(c(x_\pt))} p(\bx_\pt|\phi)\\
  &=\sum_{\bx_\pt \in c^{-1}(c(x_\pt))} \prod_{\tau<t} p(\bx_\tau|\phi)\\
  &=p(x_\pt|\phi)|c^{-1}(c(x_\pt))|,
\end{align}
for the first step, the factorization $p(x_\pt|\phi)=p(x_\tm|\phi)p(x_\ptm|\phi)$ for the second, \cref{eq:multinomial} for the third, and the fact that $c(x_\pt)=c(x_\ptm)+\delta_{x_\tm}$ for the fourth.
Finally, we combine this with the transfer entropy term which remains the same as in \cref{eq:ontrapwte} to get our main result for $\ntic^1_t$:
\begin{align}
  \ntic^1_t(x_\pt,\xi_0,\phi)&=\ntic^1_t(x_\pt)\\
  &=\log \frac{c(x_\pt)_{x_\tm}}{t}.\label{eq:iidpwonticresult}
\end{align}
Which only depends on the data sequence $x_\pt$ and neither on $\xi_0$ nor on $\phi$. It turns out to be the logarithm of the relative frequency of the last datum $x_\tm$ in the sequence $x_\pt$.

\subsection{Marginal surprise}
Here we calculate the marginal surprise of an observation $x_t$. This is the negative log probability of $x_t$ under the predictive posterior distribution. By construction of the hyperparameter process we have
\begin{align}
  q(x_t|x_\pt,\xi_0)&=q(x_t|\xi_t(x_\pt,\xi_0))\\
  &=q(x_t|\xi_0+c(x_\pt))\\
&= \int q(x_t|\hphi)q(\hphi|\xi_0+c(x_\pt)) \diff \hphi\\
&= \frac{(\xi_0+c(x_\pt))_{x_t}}{|\xi_0+c(x_\pt)|}.\label{eq:msurprise}
\end{align}
Where we used \cref{eq:iidxiupdate}
and overloaded notation and wrote $|\xi|=\sum_x (\xi)_x$. The negative logarithm of this is the marginal surprise.
\subsection{IID information gain}
Here we calculate the information gain of the hyperparameter over time. Similar to the case of NTIC this can be done for the full past $x_\pt$ and for a single observation/datum $x_\pt$.

\subsubsection{Full past information gain}
The full past information gain is defined as:
\begin{align}
  \IG_t(x_\pt,\xi_0):&=\KL[q(\hPhi|x_\pt,\xi_0)||q(\hPhi|\xi_0)]\\
&=\KL[q(\hPhi|\xi_0+c(x_\pt))||q(\hPhi|\xi_0)].
\end{align}
The KL divergence is
\begin{align}
\begin{split}
  &\KL[q(\hPhi|\xi_0+c(x_\pt))||q(\hPhi|\xi_0)]\\
  &= \int q(\hphi|\xi_0+c(x_\pt)) \log \frac{q(\hphi|\xi_0+c(x_\pt))}{q(\hphi|\xi_0)}\diff \hphi.
\end{split}
\end{align}
and a Dirichlet distribution for parameter $\xi$ is defined by
\begin{align}
  \frac{\Gamma(|\xi|)}{\prod_x \Gamma((\xi)_x)} \prod_x \hphi_x^{(\xi)_x-1}
\end{align}
where we again wrote $|\xi|=\sum_x (\xi)_x$. Here $\Gamma$ is the Gamma function. For our purposes it is sufficient to know that $\Gamma(z+1)=z\Gamma(z)$ when $n \in \Nat$.
We focus on the fraction in the logarithm in the KL divergence and plug in the definition of Dirichlet distributions:
\begin{align}
\begin{split}
  &\frac{q(\hphi|\xi_0+c(x_\pt))}{q(\hphi|\xi_0)}\\
\\
&=\frac{\Gamma(|\xi_0|+|c(x_\pt)|)\prod_x \Gamma((\xi_0)_x)}{\prod_x \Gamma((\xi_0)_x+c(x_\pt)_x) \Gamma(|\xi_0|)} \prod_x \hphi_x^{c(x_\pt)_x}\end{split}\\
&=:g(x_\pt,\xi_0) q(x_\pt|\hphi).\label{eq:qratio}
\end{align}
This means the full past information gain is  
\begin{align}
\begin{split}
 &\IG(x_\pt,\xi_0)\\
  &= \int q(\hphi|\xi_0+c(x_\pt)) \log (g(x_\pt,\xi_0) q(x_\pt|\hphi))\diff \hphi.\label{eq:fullpastig}
\end{split}
\end{align}

\subsubsection{One-step information gain}
To get the one-step information gain in at time $t$ i.e.\ in response to the observation $x_\tm$ at $\tm$ due to data sequence $x_\pt$ we plug in the hyperparameter $\xi_\tm(x_\ptm,\xi_0)=\xi_0+c(x_\ptm)$ that results from $x_\ptm$ in place of $x_0$ and set the sequence of observations $x_\pt$ to just $x_\tm$ so that $c(x_\pt)=c(x_\tm)=\delta_{x_\tm}$. This gives us the one-step information gain that occurs from time $\tm$ to time $t$ due to the last observation in observation sequence $x_\pt$. Accordingly, we define:
\begin{align}
  \begin{split}
&\IG^1_t(x_\pt,\xi_0)\\
&:=
\KL[q(\hPhi|\xi_\tm(x_\ptm,\xi_0)+\delta_{x_\tm})||q(\hPhi|\xi_\tm(x_\ptm,\xi_0))].                                                                                                              \end{split}
\end{align}
If we drop the dependence of $\xi_\tm(x_\ptm,\xi_0)$ on $(x_\ptm,\xi_0)$ to save space for the moment the function $g$ from \cref{eq:qratio} becomes:
\begin{align}
  \begin{split}
&g(x_\tm,\xi_\tm)\\
 \\
&=\frac{|\xi_\tm|\Gamma(|\xi_\tm|)\prod_x \Gamma((\xi_\tm)_x)}{(\xi_\tm)_{x_\tm}\Gamma((\xi_\tm)_{x_\tm})\prod_{x\neq x_\tm} \Gamma((\xi_\tm)_x) \Gamma(|\xi_\tm|)} \\  
  &=\frac{|\xi_\tm|}{(\xi_\tm)_{x_\tm} }\end{split}
\end{align}
Plug this into the KL-divergence
\begin{align}
\begin{split}
  &\KL[q(\hPhi|\xi_\tm+\delta_{x_\tm})||q(\hPhi|\xi_\tm)]\\
  &= \int q(\hphi|\xi_t) \log \left(\frac{|\xi_\tm|}{(\xi_\tm)_{x_\tm} } q(x_\tm|\hphi)\right)\diff \hphi\end{split}\\
  &= \Ex_{\xi_t}[\log q(x_\tm|\hPhi)]+  \log \frac{|\xi_\tm|}{(\xi_\tm)_{x_\tm} }.
\end{align}
We note here that the expectation value has a closed form solution: 
\begin{align}
  \Ex_{\xi}[\log q(x|\hPhi)]=\Psi((\xi)_x)-\Psi(|\xi|)
\end{align}
with $\Psi$ the Digamma function. 
We can then write the one-step information gain at time $t$ for given data sequence $x_\pt$ and initial hyperparameter $\xi_0$ either with the expectation value or the Digamma function:
\begin{align}
\begin{split}
  &\IG^1_t(x_\pt,\xi_0)\\
&=\Psi((\xi_t(x_\pt,\xi_0))_{x_\tm})+\\
  &\phantom{=\Psi}-\Psi(|\xi_t(x_\pt,\xi_0)|)+\\
  &\phantom{=\Psi-\Psi}-  \log \frac{(\xi_0+c(x_\ptm))_{x_\tm} }{\sum_x (\xi_0+c(x_\ptm))_x}                                                                                                                                                          
\end{split}
\end{align}
Note that the logarithm term is equal to the logarithm of the posterior predictive distribution after observations $x_\ptm$ \cite{minka2003bayesian}
\begin{align}
  q(x_\tm|\xi_0,x_\ptm)=\frac{(\xi_0+c(x_\ptm))_{x_\tm}}{\sum_x (\xi_0+c(x_\ptm))_x}.
\end{align}
and has similarities to the one-step pointwise NTIC result of \cref{eq:iidpwonticresult}. 
This can be made a bit more explicit by writing:
\begin{align}
  \frac{(\xi_0+c(x_\ptm))_{x_\tm}}{\sum_x (\xi_0+c(x_\ptm))_x}=\frac{(\xi_0)_{x_\tm}+c(x_\pt)_{x_\tm}-1}{|\xi_0|+|c(x_\pt)|-1}. 
\end{align}
With this we get two expressions for the information gain:
\begin{align}
\begin{split}
  &\IG^1_t(x_\pt,\xi_0)\\
  &= \Ex_{\xi_0+c(x_\pt)}[\log q(x_\tm|\hPhi)]+ \\ 
  &\phantom{=\Ex_{\xi_t}}- \log q(x_\tm|\xi_0,x_\ptm)
\end{split}\\
  \begin{split}
&=\Psi((\xi_0+c(x_\pt))_{x_\tm})+\\
  &\phantom{=\Psi}-\Psi(|\xi_0+c(x_\pt)|)+\\
  &\phantom{=\Psi-\Psi}-  \log \frac{(\xi_0)_{x_\tm}-1+c(x_\pt)_{x_\tm}}{|\xi_0|-1+|c(x_\pt)|}                                                                                                                                                        
\end{split}\label{eq:iidpwosurpriseresult}
\end{align}

\bibliographystyle{IEEEtran}
\bibliography{IEEEabrv,../../../bibliography}

\begin{thebibliography}{10}
\providecommand{\url}[1]{#1}
\csname url@samestyle\endcsname
\providecommand{\newblock}{\relax}
\providecommand{\bibinfo}[2]{#2}
\providecommand{\BIBentrySTDinterwordspacing}{\spaceskip=0pt\relax}
\providecommand{\BIBentryALTinterwordstretchfactor}{4}
\providecommand{\BIBentryALTinterwordspacing}{\spaceskip=\fontdimen2\font plus
\BIBentryALTinterwordstretchfactor\fontdimen3\font minus
  \fontdimen4\font\relax}
\providecommand{\BIBforeignlanguage}[2]{{%
\expandafter\ifx\csname l@#1\endcsname\relax
\typeout{** WARNING: IEEEtran.bst: No hyphenation pattern has been}%
\typeout{** loaded for the language `#1'. Using the pattern for}%
\typeout{** the default language instead.}%
\else
\language=\csname l@#1\endcsname
\fi
#2}}
\providecommand{\BIBdecl}{\relax}
\BIBdecl

\bibitem{bertschinger_information_2006}
N.~Bertschinger, E.~Olbrich, N.~Ay, and J.~Jost, ``Information and closure in
  systems theory,'' in \emph{Proceedings of the 7th German Workshop on
  Artificial Life}, Jena, 2006, pp. 9--19.

\bibitem{beer_characterizing_2014}
\BIBentryALTinterwordspacing
R.~D. Beer, ``Characterizing {autopoiesis} in the {game} of {life},''
  \emph{Artificial Life}, vol.~21, no.~1, pp. 1--19, Aug. 2014. [Online].
  Available: \url{http://dx.doi.org/10.1162/ARTL_a_00143}
\BIBentrySTDinterwordspacing

\bibitem{krakauer_information_2020}
\BIBentryALTinterwordspacing
D.~Krakauer, N.~Bertschinger, E.~Olbrich, J.~C. Flack, and N.~Ay,
  ``\BIBforeignlanguage{en}{The information theory of individuality},''
  \emph{\BIBforeignlanguage{en}{Theory in Biosciences}}, vol. 139, no.~2, pp.
  209--223, Jun. 2020. [Online]. Available:
  \url{https://doi.org/10.1007/s12064-020-00313-7}
\BIBentrySTDinterwordspacing

\bibitem{vakhrameev_measuring_2020}
\BIBentryALTinterwordspacing
D.~Vakhrameev, M.~Aguilera, X.~E. Barandiaran, and M.~Bedia, ``Measuring
  {Autonomy} for {Life}-{Like} {AI},'' \emph{Artificial Life Conference
  Proceedings}, vol.~32, pp. 589--591, Jul. 2020, publisher: MIT Press.
  [Online]. Available:
  \url{https://www.mitpressjournals.org/doi/abs/10.1162/isal_a_00308}
\BIBentrySTDinterwordspacing

\bibitem{guttenberg_neural_2016}
\BIBentryALTinterwordspacing
N.~Guttenberg, M.~Biehl, and R.~Kanai, ``Neural {Coarse}-{Graining}:
  {Extracting} slowly-varying latent degrees of freedom with neural networks,''
  \emph{arXiv:1609.00116 [cs]}, Sep. 2016, arXiv: 1609.00116. [Online].
  Available: \url{http://arxiv.org/abs/1609.00116}
\BIBentrySTDinterwordspacing

\bibitem{chang_information_2019}
\BIBentryALTinterwordspacing
A.~Y.~C. Chang, M.~Biehl, Y.~Yu, and R.~Kanai, ``Information {Closure} {Theory}
  of {Consciousness},'' \emph{arXiv:1909.13045 [q-bio]}, Sep. 2019, arXiv:
  1909.13045. [Online]. Available: \url{http://arxiv.org/abs/1909.13045}
\BIBentrySTDinterwordspacing

\bibitem{biehl_dynamics_2020}
M.~Biehl and R.~Kanai, ``\BIBforeignlanguage{eng}{Dynamics of a bayesian
  hyperparameter},'' 2020, to be published, accepted at IWAI 2020:1st
  International Workshop on Active Inference,.

\bibitem{friston_free_2019}
\BIBentryALTinterwordspacing
K.~Friston, ``A free energy principle for a particular physics,''
  \emph{arXiv:1906.10184 [q-bio]}, Jun. 2019, arXiv: 1906.10184. [Online].
  Available: \url{http://arxiv.org/abs/1906.10184}
\BIBentrySTDinterwordspacing

\bibitem{bishop_pattern_2006}
\BIBentryALTinterwordspacing
C.~Bishop, \emph{\BIBforeignlanguage{english}{Pattern {Recognition} and
  {Machine} {Learning}}}, ser. Information {Science} and {Statistics}.\hskip
  1em plus 0.5em minus 0.4em\relax New York: Springer-Verlag, 2006. [Online].
  Available: \url{//www.springer.com/jp/book/9780387310732}
\BIBentrySTDinterwordspacing

\bibitem{itti_bayesian_2006}
\BIBentryALTinterwordspacing
L.~Itti and P.~F. Baldi, ``Bayesian {Surprise} {Attracts} {Human}
  {Attention},'' in \emph{Advances in {Neural} {Information} {Processing}
  {Systems} 18}, Y.~Weiss, B.~Schölkopf, and J.~C. Platt, Eds.\hskip 1em plus
  0.5em minus 0.4em\relax MIT Press, 2006, pp. 547--554. [Online]. Available:
  \url{http://papers.nips.cc/paper/2822-bayesian-surprise-attracts-human-attention.pdf}
\BIBentrySTDinterwordspacing

\bibitem{minka2003bayesian}
T.~Minka, ``Bayesian inference, entropy, and the multinomial distribution,''
  \emph{Online tutorial}, 2003,
  \url{https://tminka.github.io/papers/minka-multinomial.pdf}.

\end{thebibliography}


\end{document}